\documentclass[letterpaper]{article} 
\usepackage{aaai24}  
\usepackage{times}  
\usepackage{helvet}  
\usepackage{courier}  
\usepackage[hyphens]{url}  
\usepackage{graphicx} 
\usepackage{natbib}  
\usepackage{caption} 
\usepackage{algorithm}

\usepackage{epsfig}
\usepackage{amsmath}
\usepackage{amssymb}
\usepackage{booktabs}
\usepackage{amsfonts}       
\usepackage{nicefrac}       
\usepackage{microtype}      
\usepackage{xcolor}         
\usepackage{multirow}
\usepackage{algpseudocode}
\usepackage{makecell}
\usepackage{caption}
\usepackage{subcaption}

\usepackage{newfloat}
\usepackage{listings}
\DeclareCaptionStyle{ruled}{labelfont=normalfont,labelsep=colon,strut=off} 
\floatstyle{ruled}
\newfloat{listing}{tb}{lst}{}
\floatname{listing}{Listing}
\title{G2P-DDM: Generating Sign Pose Sequence from Gloss Sequence \\ with Discrete Diffusion Model}
\author{
    Pan Xie\textsuperscript{\rm 1}, Qipeng Zhang\textsuperscript{\rm 1}, Taiyi Peng\textsuperscript{\rm 1}, Hao Tang\textsuperscript{\rm 2}\thanks{Corresponding Author}, Yao Du\textsuperscript{\rm 1}, Zexian Li\textsuperscript{\rm 1}
}
\affiliations{
    \textsuperscript{\rm 1}Beihang University,
    \textsuperscript{\rm 2}Carnegie Mellon University

}

\usepackage{bibentry}

\begin{document}

\maketitle

\begin{abstract}
The Sign Language Production (SLP) project aims to automatically translate spoken languages into sign sequences. Our approach focuses on the transformation of sign gloss sequences into their corresponding sign pose sequences (G2P). In this paper, we present a novel solution for this task by converting the continuous pose space generation problem into a discrete sequence generation problem. We introduce the Pose-VQVAE framework, which combines Variational Autoencoders (VAEs) with vector quantization to produce a discrete latent representation for continuous pose sequences. Additionally, we propose the G2P-DDM model, a discrete denoising diffusion architecture for length-varied discrete sequence data, to model the latent prior. To further enhance the quality of pose sequence generation in the discrete space, we present the CodeUnet model to leverage spatial-temporal information. Lastly, we develop a heuristic sequential clustering method to predict variable lengths of pose sequences for corresponding gloss sequences. Our results show that our model outperforms state-of-the-art G2P models on the public SLP evaluation benchmark. For more generated results, please visit our project page: \textcolor{blue}{\url{https://slpdiffusier.github.io/g2p-ddm}}
\end{abstract}

\section{Introduction}
\label{sec:intro}
Sign Language Production (SLP) is a crucial task for the Deaf community, involving the provision of continuous sign videos for spoken language sentences. Due to the distinct linguistic systems between sign languages and spoken languages~\cite{Pfau2018TheSO}, sign languages have different sign orders, making direct alignment mapping between them challenging. To address this issue, prior works first translate spoken languages into glosses\footnote{Sign glosses are minimal lexical items that match the meaning of signs and correspond to spoken language words.}, followed by generating sign pose sequences based on the gloss sequences (G2P)\cite{Saunders2020AdversarialTF,Saunders2020ProgressiveTF}. Finally, the generated sign pose sequence can optionally be used to produce a photo-realistic sign video\cite{Saunders2020EverybodySN}. As such, G2P is the crucial procedure of this task, and this paper focuses on it.

\begin{figure}[t]
\centering
\includegraphics[width=\linewidth]{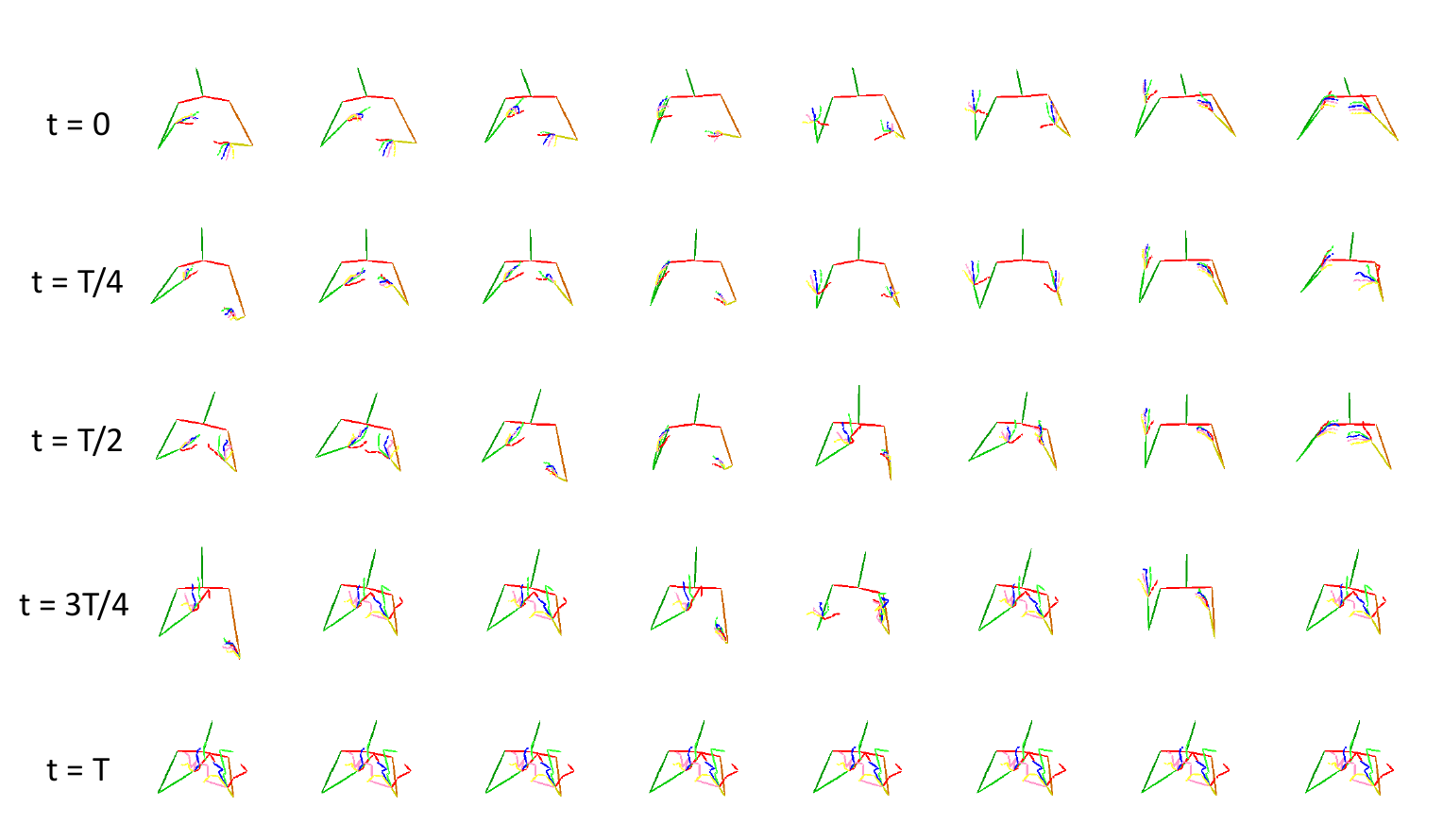}
\caption{The forward diffusion process applied to a pose sequence. The first line (t=0) represents the original pose sequence. From top to bottom (t from 0 to T), the level of noise increases gradually.} 
\label{ori_noised_pose}
\vspace{-0.4cm}
\end{figure}

Existing G2P methods can be broadly categorized as autoregressive~\cite{Saunders2020AdversarialTF,Saunders2020ProgressiveTF} or non-autoregressive~\cite{Huang2021TowardsFA}, depending on their decoding strategies. Autoregressive models generate the next pose frame based on previous frames, utilizing the teacher forcing strategy~\cite{Williams1989ALA}. However, during inference, recurrent decoding can lead to error propagation over time due to exposure bias~\cite{Schmidt2019GeneralizationIG}. To overcome this bottleneck, non-autoregressive methods have been proposed to enable the decoder to generate all target predictions simultaneously~\cite{Gu2018NonAutoregressiveNM,Ghazvininejad2019MaskPredictPD}. Huang~\textit{et al.}~\cite{Huang2021TowardsFA} introduced a non-autoregressive G2P model that generates sign pose sequences in a one-shot decoding scheme, using an External Aligner (EA) for sequence alignment learning.

Inspired by the remarkable results achieved by the recently developed Discrete Denoising Diffusion Probabilistic Models (D3PMs)~\cite{Hoogeboom2021ArgmaxFA,Austin2021StructuredDD,Gu2021VectorQD} for language and vector quantized image generation, we propose a two-stage approach in this paper. Our method involves transforming the continuous pose sequence into discrete tokens and modeling the discrete prior space using the denoising diffusion architecture. The proposed method is an iterative non-autoregressive approach that performs parallel refinement on the generated results, demonstrating expressive generative capacity.


We elaborate our approach in three steps. Firstly, we represent the pose sequence as sequential latent codes using a vector quantized variational autoencoder (VQ-VAE). Unlike image VQ-VAE~\cite{Esser2021TamingTF,Oord2017NeuralDR}, we propose a specific architecture, Pose-VQVAE, that divides the sign skeleton into three local point patches representing pose, right hand, and left hand separately. Additionally, we use a multi-codebook to maintain separated latent embedding space for each local patch, resulting in stronger feature semantics. This approach eases the difficulty in constructing mappings between the sign pose feature and the codebook feature, thus improving reconstruction quality.

Next, we present G2P-DDM, which extends the standard discrete diffusion models~\cite{Austin2021StructuredDD,Gu2021VectorQD} to model the sequential alignments between sign glosses and quantized codes of pose sequences. This approach employs a discrete diffusion model that samples the data distribution by reversing a forward diffusion process that gradually corrupts the input via a fixed Markov chain. The corruption process, depicted in Figure~\ref{ori_noised_pose}, achieved by adding noise data (\textit{e.g.}, [MASK] token), draws our attention to the mask-based generative model, Mask-Predict~\cite{Ghazvininejad2019MaskPredictPD}, which has been shown to be a variant of the diffusion model~\cite{Austin2021StructuredDD}. We explore two variants of the diffusion model for variable-length sequence generation. To better leverage the spatial and temporal information of the quantized pose sequences, we introduce a new architecture, CodeUnet, which is a "fully transformer network" designed for discrete tokens. Through iterative refinements and improved spatial-temporal modeling, our model achieves a higher quality of conditional pose sequence generation.


Finally, we address the challenging task of length prediction in non-autoregressive G2P models, as the corresponding lengths of different sign glosses are variable. To tackle this issue, we propose a novel clustering method for this specific sequential data that local adjacent frames should belong to a cluster. Taking advantage of the meaningful learned codes in the first stage, we apply the k-nearest-neighbor-based density peaks clustering algorithm~\cite{Du2016StudyOD,Zeng2022NotAT} to locate peaks with higher local density. We then design a heuristic algorithm to find the boundary between two peaks based on their semantic distance with the two peak codes. Finally, we leverage the length of each gloss as additional supervised information to predict the length of the gloss sequence during inference.

Our proposed model demonstrates significant improvement in the generation quality on the challenging RWTH-PHOENIX-WEATHER-2014T~\cite{Camgz2018NeuralSL} dataset. The evaluation of conditional sequential generation is performed using a back-translated model. Extensive experiments show that our model increases the WER score from 82.01$\%$\cite{Huang2021TowardsFA} to 77.26$\%$ for the generated pose sequence to gloss sequence, and the BLEU score from 6.66\cite{Huang2021TowardsFA} to 7.50 for the generated pose sequence to spoken language.

\begin{figure*}[t]
\centering
\includegraphics[width=0.85\linewidth]{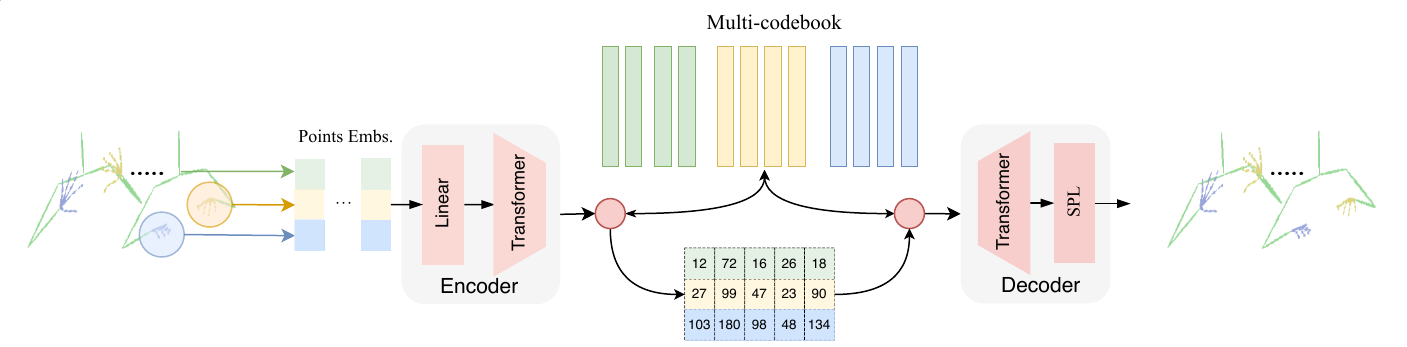}
\caption{The architecture of the first stage model Pose-VQVAE for learning the discrete latent codes.}
\label{pose_vqvae}
\vspace{-0.4cm}
\end{figure*}

\section{Related Works}

\noindent \textbf{Sign Language Production.} Most sign language works focus on sign language recognition (SLR) and translation (SLT)~\cite{Camgz2018NeuralSL,Camgz2020SignLT,Camgz2020MultichannelTF,Zhou2022SpatialTemporalMN,Xie2021PiSLTRcPS,Hu2021SignBERTPO}, aiming to translate the video-based sign language into text-based sequences. And few attempts have been made for the more challenging task of sign language production (SLP)~\cite{Stoll2018SignLP,Xiao2020SkeletonbasedCS}. Stoll~\textit{et al.} proposed the first deep SLP model, which adopts the three-step pipeline. In the core process for G2P, they learn the mapping between the sign glosses and the skeleton poses via a look-up table. After that, B. Saunders~\textit{et al.}~\cite{Saunders2020ProgressiveTF} proposed the progressive transformer to learn the mapping with an encoder-decoder architecture and generate the sign pose in an autoregressive manner in the inference. Further, B. Saunders~\textit{et al.}~\cite{Saunders2020EverybodySN} proposed a Mixture Density Network (MDN) to generate the pose sequences condition on the sign glosses and utilize a GAN-based method~\cite{Chan2019EverybodyDN} to produce the photo-realistic sign language video. B. Saunders~\textit{et al.}~\cite{SaundersCB21} translated the spoken language to sign language representation with an autoregressive transformer network and used the gloss information to provide additional supervision. Then they proposed a Mixture of Motion Primitives(MoMP) architecture to combine distinct motion primitives to produce a continuous sign language sequence. B. Saunders~\textit{et al.}~\cite{Saunders_2022_CVPR} propose a novel Frame Selection Network (FS-NET) to improve the temporal alignment of interpolated dictionary signs and SIGNGAN, a pose-conditioned human synthesis model that produces photo-realistic sign language videos direct from skeleton pose. Although they achieved state-of-the-art results, they used an additional sign language dictionary~\cite{hanke2010dgs}, meaning that each sign vocabulary has a corresponding pose sequence. Therefore, this paper did not compare their results.

Different from these methods, Huang~\textit{et al.}~\cite{Huang2021TowardsFA} proposed a non-autoregressive model to parallelly generate the sign pose sequence avoiding the error accumulation problem. They applied the monotonic alignment search~\cite{Kim2020GlowTTSAG} to generate the alignment lengths of each gloss. Our model also explores a non-autoregressive method with a diffusion strategy, and the adopted diffusion model architecture allows us to refine the results with multiple iterations.

\noindent \textbf{Discrete Diffusion Models.}  Most previous works focus on Gaussian diffusion processes that operate in continuous state spaces~\cite{Dhariwal2021DiffusionMB,Ho2020DenoisingDP,Ho2022CascadedDM,Nichol2021ImprovedDD,Rombach2021HighResolutionIS}. The discrete diffusion model is first introduced in~\cite{SohlDickstein2015DeepUL}, and it is applied to text generation in Argmax Flow~\cite{Hoogeboom2021ArgmaxFA}. To improve and extend the discrete diffusion model, D3PM~\cite{Austin2021StructuredDD} used a structured categorical corruption process to shape data generation and embed structure in the forward process. VQ-Diffusion~\cite{Gu2021VectorQD} applied the discrete diffusion model to conditional vector quantized image synthesis with a mask-and-replace diffusion strategy. Upon this work, we extend this diffusion strategy with more special states to length-varied discrete sequence data and introduce an Unet-like ``fully transformer'' network to model spatial-temporal space.

\section{The Proposed Method}

Our paper aims to improve the generation of conditional sign pose sequences through an enhanced discrete diffusion model. Our approach consists of three key components: the Pose-VQVAE for latent code learning, the G2P-DDM with CodeUnet for prior learning to generate discrete codes, and a sequential-KNN algorithm for length prediction in a non-autoregressive approach.

\subsection{Pose VQ-VAE}

In this section, we introduce how to tokenize the points of a sign pose skeleton into a set of discrete tokens. A naive approach is to treat per point as one token. However, such a points-wise reconstruction model tends to have tremendous computational cost due to the quadratic complexity of self-attention in Transformers. On the other hand, since the details of hand points are essential for sign pose understanding, treating all the points into one token leads to remarkably inferior reconstruction performance. To achieve a better trade-off between quality and speed, we propose a simple yet efficient implementation that groups the points of a sign skeleton into three local patches, representing pose, right hand, and left hand separately. Figure~\ref{pose_vqvae} illustrates the framework of our proposed Pose-VQVAE model with the following submodules. 

\noindent \textbf{Encoder.} Given a sign pose sequence of $N$ frames $\mathbf{s}=(s_1,s_2, ...,s_n, ..., s_N)\in \mathbb{R}^{N\times J \times K}$, where $\{x_n^j\}_{j=1}^J$ presents a single sign skeleton containing $J$ joints and $K$ denotes the feature dimension for human joint data. We separate these points into three local paths, $\mathbf{s}_p \in \mathbb{R}^{N\times (J_p \times K)}$, $\mathbf{s}_r \in \mathbb{R}^{N\times (J_r \times K)}$, and $\mathbf{s}_l \in \mathbb{R}^{N\times (J_l \times K)}$ for the pose, right hand, and left hand, respectively, where $J = J_p + J_r + J_l$. In the encoder module ${E}(e|\mathbf{s})$, we first transform these three point sequences into feature sequences by simple three linear layers and concatenate them together. Then we apply a spatial-temporal Transformer network to learn the long-range interactions within the sequential point features. Finally, we arrive at the encoded features ${\{e_n \in \mathbb{R}^{3\times h}\}}_{n=1}^N$. 

\noindent \textbf{Multi-Codebook.} Similar to image VQ-VAE~\cite{Oord2017NeuralDR}, we take the encoded features as inputs and convert them into discrete tokens. Specifically, we perform the nearest neighbors method $\mathcal{Q}(z|e)$ to quantize the point feature to the quantized features ${\{z_n \in \mathbb{R}^{3\times h}\}}_{n=1}^N$. The quantized features are maintained by three separate codebooks, where each codebook is of size $V$.

\noindent \textbf{Decoder.} The decoder ${D}(\tilde {\mathbf{s}}|z)$ receives the quantized features as inputs and also applies spatial-temporal Transformer to get the output features ${\{o_n \in \mathbb{R}^{3\times h}\}}_{n=1}^N$. Finally, we separate the output feature for three sub-skeleton and utilize a structured prediction layer (SPL)~\cite{Aksan2019StructuredPH} $\mathcal{P}(\tilde{s}|o)$ to reconstruct the corresponding sub-skeleton $\tilde {\mathbf{s}}_p\in \mathbb{R}^{N\times (J_r \times K)}$, $\tilde {\mathbf{s}}_l\in \mathbb{R}^{N\times (J_r \times K)}$, and $\tilde {\mathbf{s}}_r\in \mathbb{R}^{N\times (J_r \times K)}$. We adopt the SPL to rebuild the skeleton from feature because it explicitly models the spatial structure of the human skeleton and the spatial dependencies between joints. The hierarchy chains of the pose, right hand, and left hand skeleton are given in the Appendix.

\noindent \textbf{Training.} The encoder $E(e|\mathbf{s})$, tokenizer $\mathcal{Q}(z|e)$, and decoder ${D}(\tilde {\mathbf{s}}|z)$ can be trained end-to-end via the following loss function:

\begin{equation}
\begin{aligned}
\mathcal{L}_{\text{Pose-VQVAE}} = & ||\mathbf{s}_p - \tilde{\mathbf{s}}_p|| + ||\mathbf{s}_r - \tilde{\mathbf{s}}_r||  + ||\mathbf{s}_l - \tilde{\mathbf{s}}_l|| + \\
& ||sg[e] - z|| + \beta ||sg[z] - e||,
\end{aligned}
\label{eqn:equation}
\end{equation}
where $sg[\cdot]$ stands for stop-gradient operation.

\begin{figure*}[t]
\centering
\includegraphics[width=0.85\linewidth]{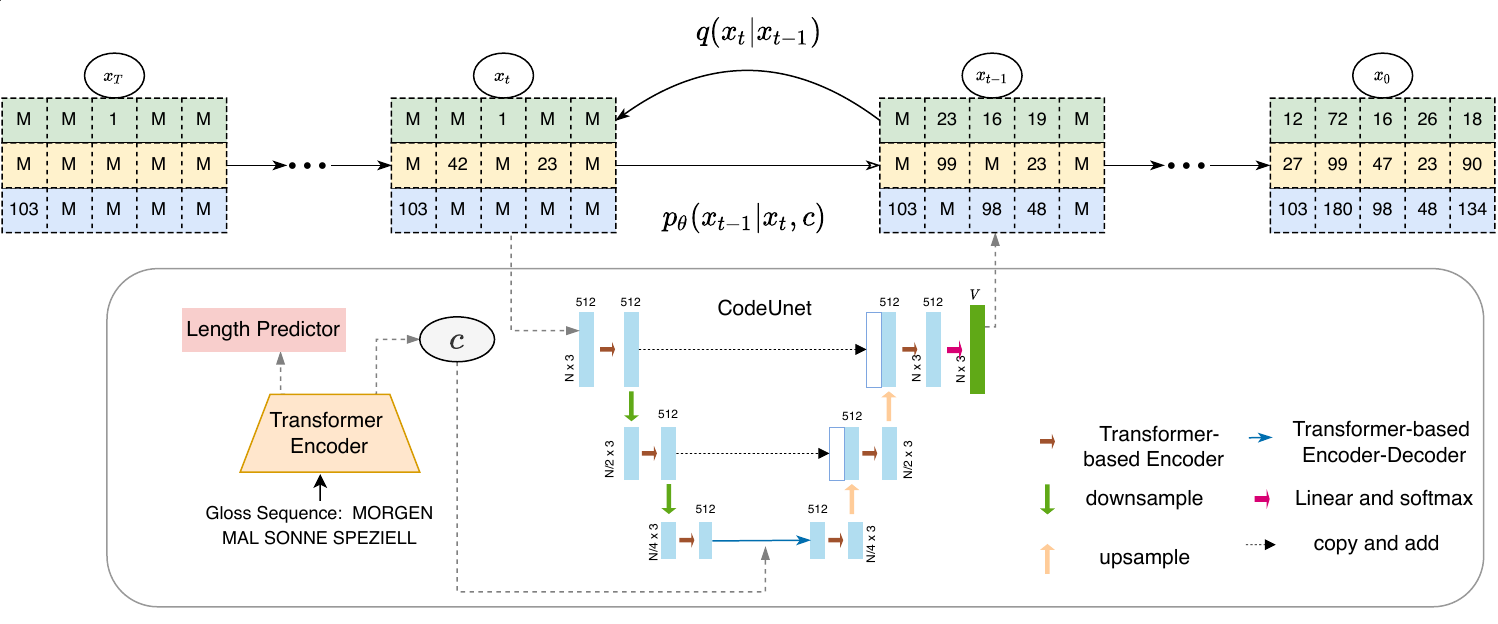}
\caption{Our approach uses a discrete diffusion model to represent the conditional sign pose sequence generation. Specifically, each quantized code is randomly masked or replaced, and a CodeUnet model is trained to restore the original data.}
\label{ddpm} 
\vspace{-0.4cm}
\end{figure*}

\subsection{G2P-DDM with CodeUnet}

To allow conditional sampling, a discrete diffusion model is trained on the latent codes obtained from the Pose-VQVAE model. Figure~\ref{ddpm} shows the architecture of our proposed G2P-DDM, which aims to model the latent space in an iterative non-autoregressive manner.  

Given a sequence of latent codes ${x_0}\in \mathbb{R}^{N\times 3}$ obtained from the vector quantized model, where $x_{0}^{(i,j)} \in \{1,2,...,V\}$ at location $(i,j)$ represents the index within the codebook. The diffusion process aims to corrupt the original data $x_0$ via a fixed Markov chain $p(x_{t}|x_{t-1})$ by adding a small amount of noise continuously. After a fixed $T$ timesteps, it produces a sequence of increasingly noisy data $x_1,..,x_T$ with the same dimensions as $x_0$, and $x_T$ becomes a pure noise sample.

For the scalar discrete variables with $V$ categories $x_t^{(i,j)}\in [1, V]$, the forward transition probabilities from $x_{t-1}$ to $x_{t}$ can be represented by matrices $[Q_t]_{mn} = q(x_t=m|x_{t-1}=n) \in \mathbb{R}^{V\times V}$. Note that we omit the superscripts $(i,j)$ to avoid confusion. Then the forward diffusion process can be written as:

\begin{equation}
q(x_t|x_{t-1})=\mathbf{x}_t^TQ_t\mathbf{x}_{t-1},
\end{equation}

\noindent where $\mathbf{x}_t\in \mathbb{R}^{V\times 1}$ is the one-hot version of $x_t$ and $Q_t\mathbf{x}_{t-1}$ is the categorical distribution for $x_t$. A nice property of the above Markov diffusion process is that we can sample $x_t$ as any timestep directly from $x_0$ as:

\begin{equation}
q(x_t|x_0)=\mathbf{x}_t^T\bar{Q}_t\mathbf{x}_{0}, \text{with } \bar{Q}_t=Q_t\dots Q_1.
\label{eqn3}
\end{equation}

D3PM~\cite{Austin2021StructuredDD} formulates the transition matrix $Q_t\in \mathbb{R}^{V\times V}$ by introducing a small number of uniform noises to the categorical distribution. Based on D3PM, VQ-Diffusion~\cite{Gu2021VectorQD} proposes a mask-and-replace diffusion strategy that not only replaces the previous value but also inserts [MASK] token to explicitly figure out the tokens that have been replaced. We extend this mask-and-replace strategy to our variable-length sequence modeling. Since the length of pose sequences may be different in a minibatch, we have to add two special tokens, [MASK] and [PAD] tokens, so each token has $V+2$ states. The mask-and-replace diffusion process can be defined as follows: each token has a probability of $\alpha_t$ to be unchanged, $V\beta_t$ to be uniformly resampled, and $\gamma_t=1-\alpha_t-V\beta_t$ to be replaced with [MASK] token. Note that [MASK] and [PAD] tokens always keep their own state. The transition matrix $Q_t\in \mathbb{R}^{(V+2)\times (V+2)}$ is formulated as the second matrix of the following:
\vspace{-0.2cm}
\begin{equation}
Q_t = \small{\begin{bmatrix}
\alpha_t + \beta_t & \beta_t & \cdots & \beta_t  & 0 & 0 \\ 
\beta_t & \alpha_t + \beta_t & \cdots & \beta_t & 0 & 0 \\ 
\vdots & \vdots & \ddots & \vdots & \vdots & \vdots \\
\beta_t & \beta_t & \cdots & \alpha_t + \beta_t  & 0 & 0 \\
\gamma_t & \gamma_t & \cdots & \gamma_t  & 1 & 0 \\
0 & 0 & \cdots & 0  & 0 & 1 \\
\end{bmatrix}.}
\end{equation}

Finally, the categorical distribution of $\mathbf{x}_t$ can be derived as following using reparameterization trick:
\begin{equation}
\begin{aligned}
\text{when} \,  x_0 \ne V+2 ,\quad
& \bar{Q_t}\mathbf{x}_0 = \begin{cases} \bar{\alpha}_t + \bar{\beta}_t, & x_t=x_0 \\ 
\bar{\beta}_t, & x_t \ne x_0 \text{ and } x_t \leq V \\
\bar{\gamma}_t, & x_t = V+1 \\
0, & x_t = V+2 \\
\end{cases} \\
\text{when} \,  x_0 = V+2 ,\quad
& \bar{Q_t}\mathbf{x}_0 = \begin{cases} 0, & x_t \ne V+2 \\
1, & x_t = V+2 \\
\end{cases}
\end{aligned}
\label{eqn5}
\end{equation}
where $\bar{\alpha}_t=\prod_{i=1}^t\alpha_i$, $\bar{\gamma}_t=1-\prod_{i=1}^t(1-\gamma_i)$, and $\bar{\beta}_t=(1-\bar{\alpha}_t-\bar{\gamma}_t)/V$. Therefore, we can directly sample $x_t$ within the computation cost $O(V)$. A visualized example of the diffusion process is shown in Figure~\ref{ori_noised_pose}, we first get the noised latent codes by $q(x_t|x_t)$ and decode them to sign skeleton with Pose-VQVAE decoder module.

The reverse denoising process is similar to D3PM~\cite{Austin2021StructuredDD} and VQ-Diffusion~\cite{Gu2021VectorQD}. The relevant derivation process is given in the appendix.

\noindent \textbf{CodeUnet for Model Learning.} Most image diffusion models~\cite{Dhariwal2021DiffusionMB,Ho2020DenoisingDP,Song2021ScoreBasedGM} adopt the Unet~\cite{Ronneberger2015UNetCN} as their architectures since it is effective for data with spatial structure. However, directly applying the Unet in discrete sequence generation, \textit{e.g.}, text generation~\cite{Austin2021StructuredDD} and quantized image synthesis~\cite{Gu2021VectorQD}, will bring information leakage problem since the convolution layer over adjacent tokens may provide shortcuts for the mask-based prediction~\cite{Nawrot2021HierarchicalTA}. Therefore, Austin~\textit{et al.}~\cite{Austin2021StructuredDD} and Gu~\textit{et al}~\cite{Gu2021VectorQD} used the token-wise Transformer framework to learn the distribution $p_{\theta}(\tilde{x}_0|x_t,c)$. In this work, to incorporate the advantages of Unet and Transformer networks, we propose a novel architecture, CodeUnet, to learn the spatial-temporal interaction for our quantized pose sequence generation.

As shown in Figure~\ref{ddpm}, the CodeUnet consists of a contracting path (left side), an expansive path (right side), and a middle module. The middle module is an encoder-decoder Transformer framework. The encoder consists of 6 Transformer blocks. It takes the gloss sentence as input and obtains a conditional feature sequence. The decoder has two blocks. Each block has a self-attention, a cross-attention, a feed-forward network, and an Adaptive Layer Normalization~(AdaLN) \cite{Ba2016LayerN,Gu2021VectorQD}. The AdaLN operator is devised to incorporate timestep $t$ information as $\text{AdaLN}(h,t)=\alpha_t\text{LayerNorm}(h) + \beta_t$, where $h$ is the intermediate activations, $\alpha_t$ and $\beta_t$ are obtained from a linear projection of the timestep embedding.

The contracting and expansive paths are hierarchical structures, and each level has two Transformer encoder blocks. For downsampling in contracting path, given the feature of quantized pose sequence, \textit{e.g.}, $h\in \mathbb{R}^{N\times 3\times d_{\text{model}}}$, where $d_{\text{model}}$ is the feature dimension, we first sample uniformly with stride $2$ in the temporal dimension and remain constant in the spatial dimension. Then we set the downsampled feature as query $Q\in \mathbb{R}^{N/2\times 3\times d_{\text{model}}}$, and keep key $K$ and value $V$ unchanged for the following attention network. In the upsampling of the expansive path, we directly repeat the feature $2$ times as a query, but the key and value remain for the following attention network:
\begin{equation}
\begin{aligned}
\forall n=1,...,N, Q^{\text{up}}_n = h_{n//2}, K^{\text{up}} = V^{\text{up}} = h,
\end{aligned}
\end{equation}
where $\cdot//\cdot$ denotes floor division. Finally, a linear layer and a softmax layer are applied to make the prediction.

\begin{table*}
 \renewcommand\arraystretch{1.0}
  \centering
  \begin{tabular}{c|cccccc}
  \toprule
  {\small{Method}} & {\small{WER}} & {\small{BLEU-1}} & {\small{BLEU-2}} & {\small{BLEU-3}} & {\small{BLEU-4}} & {\small{DTW-MJE}} \\
  \midrule
  $\text{\small{PTR}}^{\dag}$~\cite{Saunders2020ProgressiveTF} & \small{94.65} & \small{11.45} & \small{7.08} & \small{5.08} & \small{4.04} & \small{0.191} \\
  \hline
  \small{NAT-AT}~\cite{Huang2021TowardsFA} & \small{88.15} & \small{14.26} & \small{9.93} & \small{7.11} & \small{5.53} & \small{0.177} \\
  \small{NAT-EA}~\cite{Huang2021TowardsFA} & \small{82.01} & \small{15.12} & \small{10.45} & \small{7.99} & \small{6.66} & \small{0.146} \\
  \hline
  \textbf{\small{G2P-AR} (Ours)} & \small{85.27} & \small{14.26} & \small{10.02} & \small{7.57} & \small{5.94} & \small{0.172} \\
  \textbf{\small{G2P-MP} (Ours)} & \small{79.38} & \small{15.43} & \small{10.69} & \small{8.26} & \small{6.98} & \small{0.146} \\
  \textbf{\small{G2P-DDM} (Ours)} & \textbf{\small{77.26}} & \textbf{\small{16.11}} & \textbf{\small{11.37}} & \textbf{\small{9.22}} & \textbf{\small{7.50}} & \textbf{\small{0.116}} \\
  \hline
   $\text{\small{GT}}^{\dag}$ & \small{55.93} & \small{24.12} & \small{16.77} & \small{12.80} & \small{10.58} & 0.0 \\
  \bottomrule
  \end{tabular}
   \caption{Quantitative results for the G2P task on the RWTH-PHOENIX-WEATHER-2014T test dataset. $\dag$ indicates the results is provided by Huang et al.~\cite{Huang2021TowardsFA}. Note that, GT refers to the validation metrics obtained by using the original pose sequence extracted from the video and then applying a back-translation method.}
  \label{tab:comparison}
\end{table*}

\subsection{Length Prediction with Sequential-KNN}
In this section, inspired by~\cite{Zeng2022NotAT}, which merges tokens with similar semantic meanings from different locations, we propose a novel clustering algorithm to get the lengths for corresponding glosses. Specifically, given a token sequence that is obtained from the Pose-VQVAE model, we compute the local density $\rho$ of each token according to its k-nearest-neighbors:
\begin{equation}
\begin{aligned}
\rho_i=exp(-\dfrac{1}{k}\sum_{z_j\in \text{KNN}(z_i)} \lVert z_i - z_j \rVert_2^2),\,\text{where}\,|i-j| <= l
\end{aligned}
\label{eqn:knn}
\end{equation}
where $i,j$ is the position in the sequence, and $l$ is a predefined hyperparameter indicating that we only consider the local region since the adjacent tokens are more likely to belong to a gloss. $z_i$ and $z_j$ are the latent feature for $i^{th}$ and $j^{th}$ tokens. $\text{KNN}(x_i)$ represents the k-nearest neighbors for $i^{th}$ token.

We assign $\{p_1, ..., p_M\}$ positions with a higher local density as the peaks, where $M$ is the length of the gloss sequence. Then between two adjacent peaks, for example, $p_1$ and $p_2$, we sequentially iterate from $p_1$ to $p_2$ and find the first position that is farther from $z_{p_1}$ and closer to $z_{p_2}$, which is the boundary we determined. After finding these boundaries, we get the lengths of the contiguous pose sequence for its corresponding glosses. As shown in Figure~\ref{ddpm}, we define the obtained lengths as $\{L_1,.., L_M\}$, and the Transformer encoder for gloss sequence is trained under the supervised information of lengths. For each gloss word, we predict a number from $[1, P]$, where $P$ is the maximum length of the target pose sequence. Mathematically, we formulate the classification loss of length prediction as:
\begin{equation}
\begin{aligned}
\mathcal{L}_{\text{len}} = \frac{\delta}{M}\sum_{i}^{M}\sum_{j}^{P}(-L_i=j)\log p(L_i|c).
\end{aligned}
\label{eqn12}
\end{equation}

In the training of the discrete diffusion mode, $\mathcal{L}_{\text{len}}$ is trained together with a coefficient $\delta$. In the inference, we predict the length of glosses, and their summation is the length of the target pose sequence.

In summary, we arrive at our proposed two-stage approach, G2P-DDM, with the first-stage Pose-VQVAE model and the second-stage discrete diffusion model with a length predictor. 

\section{Experiments}

\noindent\textbf{Datasets.} We evaluate our G2P model on RWTH-PHOENIX-WEATHER-2014T dataset~\cite{Camgz2018NeuralSL}. It is the \textit{only} publicly available SLP dataset with parallel sign language videos, gloss annotations, and spoken language translations. This corpus contains 7,096 training samples (with 1,066 different sign glosses in gloss annotations and 2,887 words in German spoken language translations), 519 validation samples, and 642 test samples.

\noindent\textbf{Evaluation Metrics.} Following the widely-used setting in SLP~\cite{Saunders2020ProgressiveTF}, we adopt the back-translation method for evaluation. Specifically, we utilize the state-of-the-art SLT~\cite{Camgz2020SignLT} model to translate the generated sign pose sequence back to gloss sequence and spoken language, where its input is modified as pose sequence. Specifically, we compute BLEU~\cite{Papineni2002BleuAM} and Word Error Rate (WER) between the back-translated spoken language translations and gloss recognition results with ground truth spoken language and gloss sequence. Although this evaluation method may introduce noise, it is currently the prevailing approach in SLP models, and we adopt it to ensure a fair comparison with existing methods.

\noindent\textbf{Data Processing.} Since the RWTH-PHOENIX-WEATHER-2014T dataset does not contain pose information, we generate the pose sequence as the ground truth. Following B. Saunders \textit{et al.}~\cite{Saunders2020ProgressiveTF}, we extract 2D joint points from sign video using OpenPose~\cite{Cao2021OpenPoseRM} and lift the 2D joints to 3D with a skeletal model estimation improvement method~\cite{Zelinka2020NeuralSL}. Finally, similar to ~\cite{Stoll2018SignLP}, we apply skeleton normalization to remove the skeleton size difference between different signers.

\noindent\textbf{Model Settings.}  The Pose-VQVAE consists of an Encoder, a Tokenizer, and a Decoder. The Encoder contains a linear layer to transform pose points to a hidden feature with a dimension set as 256, a 3-layer Transformer module with divided space-time attention~\cite{Bertasius2021IsSA}. The Tokenizer maintains a codebook with a size set as 2,048. The Decoder contains the same 3-layer Transformer module as the Encoder and an SPL layer to predict the structural sign skeleton. For the discrete diffusion model, we set the timestep $T$ as 100. All Transformer blocks of CodeUnet have $d_{\text{model}}{=}512$ and $N_{\text{depth}}{=}2$. The size of the local region $l$ in Eq.~\eqref{eqn:knn}, is set as $16$, which is the average length of a gloss. And the number of nearest neighbors $k$ is set as 16. We train the model on 8 NVIDIA Tesla V100 GPUs. We include all hyperparameters settings and the details of implementation in the Appendix.

\subsection{Comparisons with State-of-the-Art Methods}

\noindent\textbf{Competing Methods.} We compare our G2P-DDM with previous state-of-the-art G2P models. \textbf{Progressive Transformer} (PTR)~\cite{Saunders2020ProgressiveTF} is the first SLP model to tackle the G2P problem in an autoregressive manner. Since they use the ground-truth first sign pose frame and timing information, their reported results are not comparable to ours. Thus we adopt the results reported by Huang~\textit{et al.}~\cite{Huang2021TowardsFA}. \textbf{NAT-EA}~\cite{Huang2021TowardsFA} proposes a non-autoregressive method to directly predict the target pose sequence with the External Aligner (EA) to learn alignments between glosses and pose sequences. \textbf{NAT-AT} is the NAT model without EA that uses the decoder-to-encoder attention to learn the alignments. 

\noindent\textbf{Quantitative Comparison.} The comparison between our G2P-DDM and the competing methods is shown in Tabel~\ref{tab:comparison}. Note that, the evaluation results of the $\text{\small{GT}}^{\dag}$ are lower than the reported results in the state-of-the-art SLT~\cite{Camgz2020SignLT} model. This is because the evaluation results obtained using the pose sequence are inferior to those obtained using photo-realistic content~\cite{Saunders_2022_CVPR}.

The row of \textbf{G2P-AR} refers to the vector quantized model with an autoregressive decoder. The row of \textbf{G2P-MP} refs to the vector quantized model with the Mask-Predict~\cite{Ghazvininejad2019MaskPredictPD} strategy, which is also a variant of discrete diffusion model~\cite{Austin2021StructuredDD}. \textbf{G2P-DDM} refs to the vector quantized model with mask-and-replace diffusion strategy. As indicated in Table~\ref{tab:comparison}, both diffusion-based models outperform the state-of-the-art G2P models with relative improvements on WER score by 5.7$\%$ (82.01 $\rightarrow$ 77.26) and on BLEU-4 by 12.6$\%$ (6.66 $\rightarrow$ 7.50). This shows the effectiveness of the iterative mask-based non-autoregressive method on the vector quantized pose sequence. In addition, the Mask-Predict strategy is a mask-only strategy similar to G2P-DDM with $\bar{\gamma}_T=1$. Therefore, G2P-DDM achieves better performance than G2P-MP. This reflects that the mask-and-replace strategy is superior to the mask-only strategy.

\begin{figure}[t]
\centering
\includegraphics[width=0.9\linewidth]{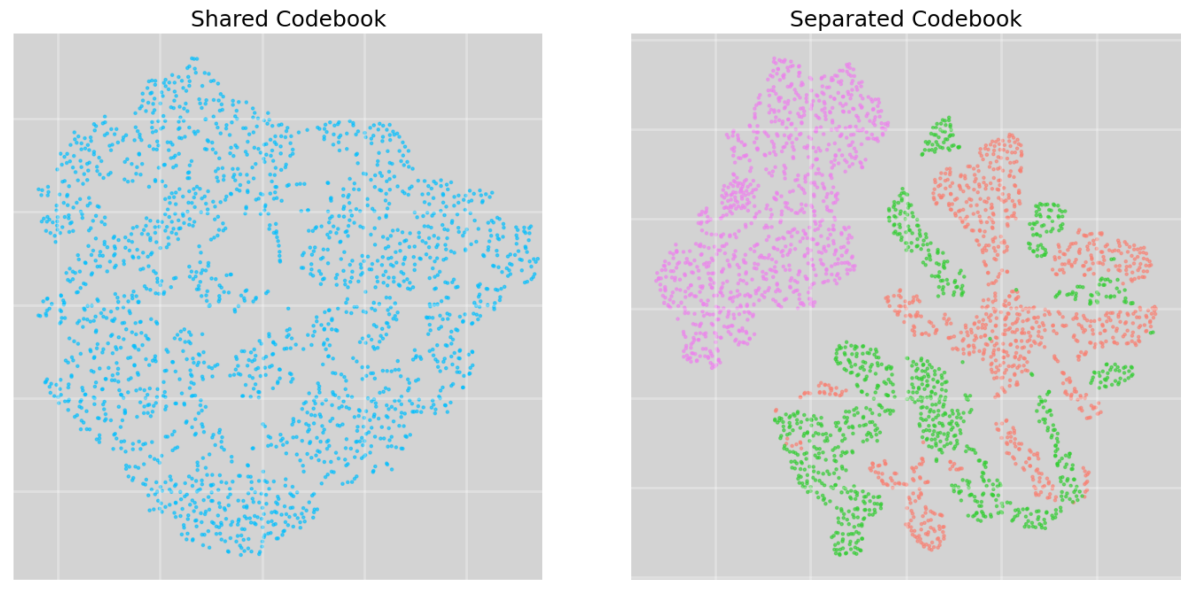}
\caption{Visulization of latent vectors in the shared codebook and separated codebooks. In the separated codebook, the pink part is for the pose, and the green and orange parts represent the left and right hands, respectively.}
\label{abl:codebook} 
\vspace{-0.4cm}
\end{figure}

\begin{figure}[!t]
\centering
\includegraphics[width=0.45\textwidth]{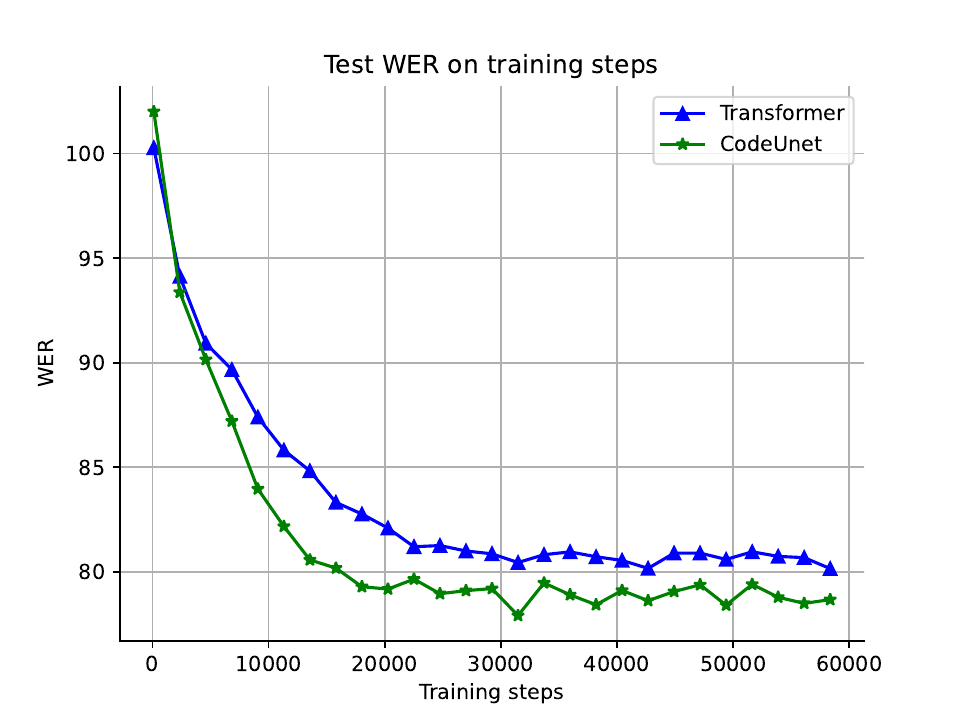}
\caption{Ablation on the design of prediction model.}
\label{abl:codeunet} 
\vspace{-0.4cm}
\end{figure}

\begin{figure*}[t]
\centering
\includegraphics[width=0.9\linewidth]{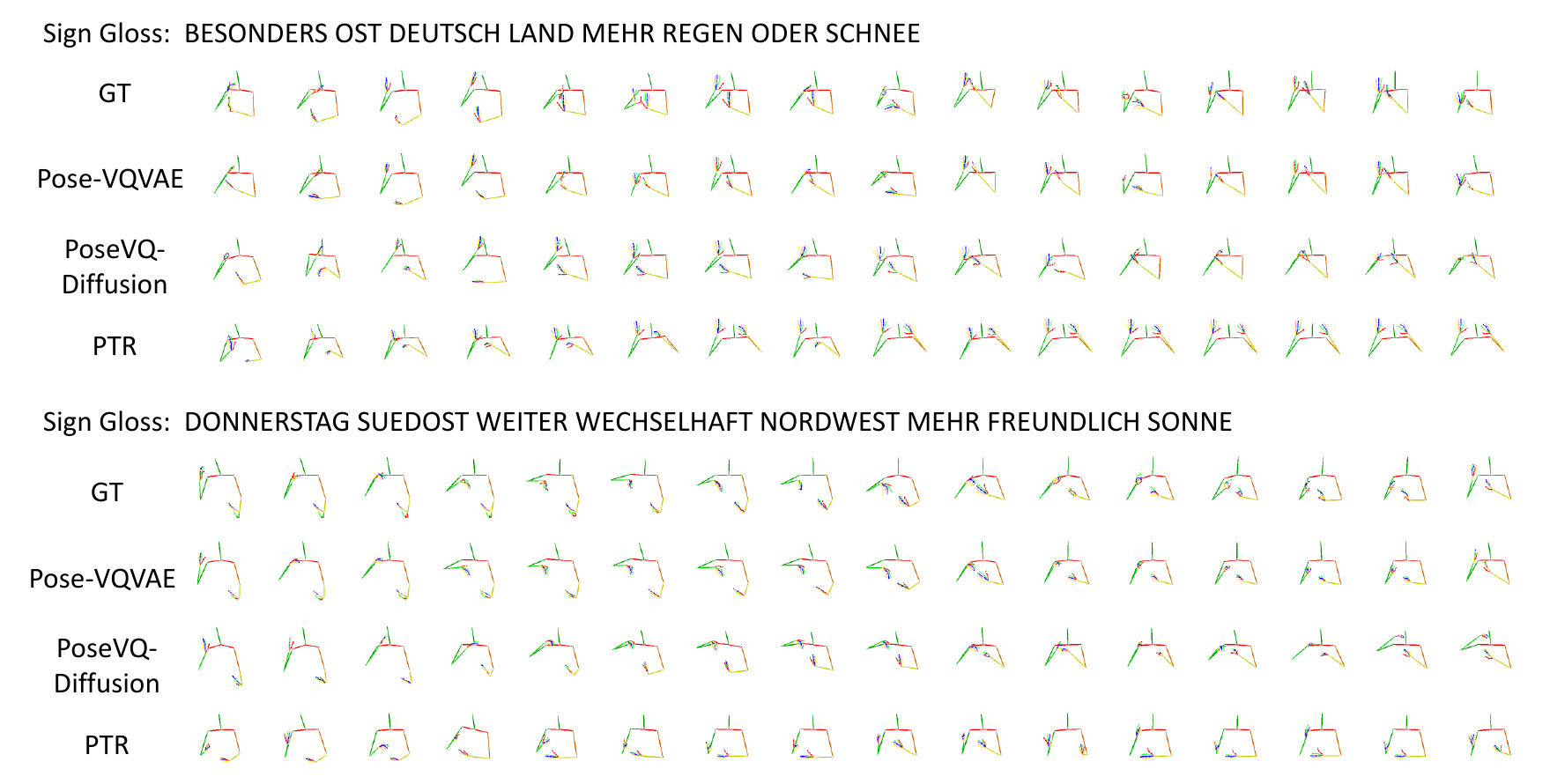}
\caption{G2P qualitative results. We show several examples of generated sign pose sequences compared with Pose-VQVAE and previous G2P model~\cite{Saunders2020ProgressiveTF}. For readability, we sampled every 5 frames for a total of 16 frames. See the Appendix for more results.}
\label{fig:results} 
\vspace{-0.4cm}
\end{figure*}

\begin{table}[t]
\centering
\smallskip
\resizebox{0.9\linewidth}{!}{
\begin{tabular}{c|c|c|c}
\hline
\small{{local patches}} & \small{{codebook (size)}} & \small{{MSE} ($\downarrow$)} & \small{{WER} ($\downarrow$)} \\
\hline
\small{joint} & \small{shared (2048)} & \small{0.0242} & - \\
\small{separated} & \small{shared (2048)} & \small{0.0139} & \small{78.21}\\
\small{separated} & \small{shared (3072)} & \small{0.0131} & \small{78.15}\\
\small{separated} & \small{separated (1024*3)} & \small{\textbf{0.0113}} & \small{\textbf{77.26}}\\
\hline
\end{tabular}}
\caption{Ablation on the design of Pose-VQVAE reconstruction model.}
\label{abl:model_design}
\vspace{-0.2cm}
\end{table}

\subsection{Model Analysis and Discussions}
We also investigate the effects of different components and design choices of our proposed model.

\noindent\textbf{Analysis of The Design of Pose-VQVAE.} 
As shown in Table~\ref{abl:model_design}, we study the design of our Pose-VQVAE model. Pose-VQVAE-joint-shared means we compress all points into one token with one shared codebook. Pose-VQVAE-separated-shared means the points are separated into three local patches according to the structure of a sign skeleton, and the latent embedding space is maintained with one shared codebook. Pose-VQVAE-separated-separated means the points are separated into three local patches, and the latent vectors are maintained with three codebooks separately.

\begin{table}
\centering
\resizebox{0.7\linewidth}{!}{
\begin{tabular}{c|c|c|c|c|c}
\hline
& \multicolumn{5}{c}{Training Steps}            \\\hline
\multirow{5}{*}{\rotatebox{90}{Infer. Steps}} &  & \small{$20$}  & \small{$50$}  & \small{$100$} & \small{$200$}  \\\cline{2-6}
& \small{$20$}   & \small{79.53}  & \small{79.40} & \small{78.25}  & \small{78.62}   \\\cline{2-6}
& \small{$50$}  & -   & \small{79.31} & \small{77.69} & \small{78.23}  \\\cline{2-6}
& \small{$100$} & -   & -   & \small{\textbf{77.26}} & \small{78.18} \\\cline{2-6}
& \small{$200$} & -   & -   & -   & \small{78.15}  \\
\hline
\end{tabular}}
\caption{Ablation on training steps and inference steps.}
\label{abl:train_infer_step}
\vspace{-0.4cm}
\end{table}

Experimental results in Table~\ref{abl:model_design} show that Pose-VQVAE-separated-separated achieves much better reconstruction (MSE) performance. This indicates that compressing all skeleton points into one token embedding is not advisable, leading to information loss. Using separated latent feature spaces for different local regions, that is, three codebooks can achieve better reconstruction quality and generation performance. To further explain this phenomenon, we visualize the latent space vectors of shared codebooks and separated codebooks with T-SNE~\cite{van2008visualizing}. As shown in Figure~\ref{abl:codebook}, the latent space vectors corresponding to the left-hand and right-hand local regions are easily confused because of their close distances. Therefore, separated codebooks can reduce the difficulty in constructing mappings between the sign pose feature and the codebook feature, thus learning better latent space and reconstruction quality. The second row of Figure~\ref{fig:results} shows the sample of sign pose sequences reconstructed by Pose-VQVAE-separated-separated.

\noindent\textbf{CodeUnet vs. Transformer.} 
For a fair comparison, we replace our CodeUnet with a Transformer network, keeping other settings the same. As shown in Figure~\ref{abl:codeunet}, the diffusion-based model with our CodeUnet achieves better performance on the back-translate evaluation. This phenomenon suggests that the hierarchical structure of CodeUnet makes it particularly effective for data with spatial structure. Moreover, the curve in the figure shows that CodeUnet coverages faster than Transformer. Having said that, due to sign pose sequences being temporally redundant, the compression of CodeUnet in the time dimension makes it more efficient in training.

\noindent\textbf{Number of Timesteps.} 
We compare the performance of the model with different numbers of training steps. As shown in the left two columns of Table~\ref{abl:train_infer_step}, we find that the results get better when the training step size is increased from 20 to 100. As it increased further, the results seemed to saturate. Therefore, we set the training step to 100 to trade off performance and speed. Besides, within the same training steps, increasing the number of inference steps yields better results.

\noindent\textbf{Deaf User Evaluation} In our final user evaluation, we provided 50 pose sequences generated by our proposed method and a baseline method~\cite{Saunders2020ProgressiveTF}, and asked 7 participants to compare which one was closer to the ground truth pose sequence. The results showed that 319/350 preferred our method, while only 31/350 chose the baseline method. This clearly demonstrates the superiority of our proposed approach.



\section{Conclusion}

We present a novel paradigm for text-based sign pose sequence generation. Specifically, we first devise a specific architecture Pose-VQVAE with a multi-codebook to learn semantic discrete codes by reconstruction. Then we extend the discrete diffusion method with special states to model the alignments between sign glosses and length-varied quantized code sequences. Further, a ``fully transformer'' network CodeUnet is proposed to model the spatial-temporal information in discrete space. Finally, we propose a sequential-KNN algorithm to learn the length of corresponding glosses and then predict the length as a classification task. Our extensive experiments show that our proposed G2P-DDM framework outperforms previous state-of-the-art methods.

\bibliography{aaai24}

\end{document}